\documentclass{article}

\usepackage{arxiv}

\usepackage[utf8]{inputenc}
\usepackage[T1]{fontenc}
\usepackage{graphicx}
\usepackage{amsmath}
\usepackage{amsfonts}
\usepackage{textcomp}
\usepackage{url}
\usepackage[authoryear]{natbib}
\setcitestyle{aysep={}}
\usepackage{hyperref}

\usepackage{seqsplit}
\usepackage{xcolor}
\usepackage{siunitx}
\usepackage{mathtools}
\usepackage{amssymb}
\usepackage{bm}
\usepackage{multirow}
\usepackage{booktabs}
\newcommand{\botrule}{\bottomrule}% sn-jnl table rule
\definecolor{toolyellow}{HTML}{D4A000}
\definecolor{frontpink}{HTML}{C71585}
\definecolor{pistonviolet}{HTML}{8A2BE2}
\definecolor{pistonlime}{HTML}{228B22}
\definecolor{screwcyan}{HTML}{008B8B}
\definecolor{tubegreen}{HTML}{4CAF50}
\definecolor{rearblue}{HTML}{00008B}
\definecolor{bottomorange}{HTML}{E65100}
\definecolor{assembliesgreen}{HTML}{008080}
\definecolor{rodnutpink}{HTML}{FF1493}
\newcommand{\articletitle}{A Vision-Based Framework for Manual Assembly Verification and Adaptive Operator Assistance}

\hypersetup{
    pdftitle={\articletitle},
    pdfauthor={Naveen Joseph Delphin},
    pdfsubject={Vision-based manual assembly monitoring and guidance},
    pdfkeywords={manual assembly, computer vision, adaptive assistance, finite-state machines, object detection, human-centered manufacturing},
    colorlinks=false,
    pdfborder={0 0 0}
}

\title{\articletitle}

\author{
  Naveen Joseph Delphin\thanks{Corresponding author.} \\
  \normalfont Fraunhofer Italia Research, Bolzano, Italy \\
  \normalfont University of Bologna, Bologna, Italy \\
  \normalfont\texttt{naveenjosephdelphin@gmail.com}
  \And
  Leonardo Venturoso \\
  \normalfont Fraunhofer Italia Research \\
  \normalfont Bolzano, Italy \\
  \normalfont\texttt{leonardo.venturoso@fraunhofer.it}
  \AND
  Isabella Soraruf \\
  \normalfont Fraunhofer Italia Research \\
  \normalfont Bolzano, Italy \\
  \normalfont\texttt{isabella.soraruf@fraunhofer.it}
  \And
  Marco Todescato \\
  \normalfont Fraunhofer Italia Research \\
  \normalfont Bolzano, Italy \\
  \normalfont\texttt{marco.todescato@fraunhofer.it}
}

\newcommand{\abstracttext}{Manual assembly remains difficult to monitor because operator actions, components, and task progress must be interpreted together. This paper presents a modular framework based on a single overhead red-green-blue (RGB) camera, combining object detection, hand-motion tracking, region-aware interaction detection, finite-state verification, task-specific pose classification, and configurable visual guidance. The system was evaluated during pneumatic-piston assembly and disassembly with eight participants under three assistance conditions: full, mid, and no guidance. Mixed-effects analysis identified an association between assistance condition and a composite measure of motion quality, with lower estimated scores under mid and no guidance relative to full guidance. However, the condition-by-trial interaction was not statistically significant, and the estimated differences in learning trajectories remain descriptive. No condition differences were detected in completion time or failed-stage counts. The study demonstrates the integration of visual perception, procedural verification, and configurable operator feedback in a single-camera workflow. Selecting assistance according to operator expertise remains a design objective for future investigation.}

\begin{document}

\maketitle

\begin{abstract}
\abstracttext
\end{abstract}


%%=============================================================%%
%% 1. INTRODUCTION                                             %%
%%=============================================================%%
\section{Introduction}\label{sec:introduction}

Human-centered manufacturing requires monitoring that relates operator actions to parts, workspace locations, and procedural state, rather than recognizing objects or movements in isolation \citep{doi/10.2777/308407, zhang2025stage, zhou2024assembly}. For manual assembly, a system must infer whether an interaction is occurring, determine which component is involved and whether its task-relevant orientation is acceptable, and evaluate the action in the context of the prescribed sequence. These inferences are difficult under occlusion, visually similar parts, and reflective surfaces.

Prior work addresses components of this problem, including object and action recognition, trajectory-based verification, and procedural monitoring \citep{aganian2023object, kawase_skill_improvement, zhang2025stage, zhou2024assembly}. The engineering question addressed here is how to connect visual evidence to explicit procedural reasoning and operator-facing feedback in a single-camera RGB system. The contribution is an integrated, replaceable pipeline: learned perception provides detections, hand landmarks, and task-specific orientation classes; region-based interaction logic and two finite-state machines use these outputs to verify manipulations and assembly stages; and a configurable interface presents one of three guidance levels. The work does not introduce a new general-purpose detection or pose-estimation algorithm.

We evaluate the system in a controlled laboratory study in which eight participants assembled and disassembled a pneumatic piston under full, mid, and no guidance. The study asks whether guidance condition is associated with motion quality and whether the condition-by-trial relationship differs over repeated practice. The primary contribution is this system integration and within-participant comparison. The laboratory task provides an initial evaluation of the integrated workflow, but does not establish performance in production settings or across tasks, workstations, and camera views.

The remainder of this article is organized as follows.
Section~\ref{sec:related-work} reviews relevant work on assembly monitoring, action recognition, object detection, and pose estimation. Section~\ref{sec:methodology} presents the proposed system.
The experimental methodology is described in Section~\ref{sec:experimental-setup}, while the results are reported in Section~\ref{sec:results}. Section~\ref{sec:discussion} discusses the main findings and limitations, and Section~\ref{sec:conclusion} concludes the article.

%%=============================================================%%
%% 2. RELATED WORK                                             %%
%%=============================================================%%
\section{Related Work}\label{sec:related-work}

Wearable sensors, inertial measurement units (IMUs), force sensors, depth cameras, and RGB vision systems have been used to identify assembly operations and detect errors in manual procedures \citep{aganian2023object, kawase_skill_improvement, zhang2025stage, zhou2024assembly}. Vision-based approaches can observe operators, tools, and components without attaching sensors to the operator \citep{Quadrini}, although camera-based inference remains sensitive to occlusion, viewpoint, and appearance. The present system combines spatial and temporal visual cues in a single-camera RGB pipeline and evaluates their use within procedural verification and configurable guidance.

For action recognition, many modern methods use deep learning models based on RGB frames, skeletons, or multimodal data \citep{Sun_2022}. Object-aware skeleton methods improve recognition over skeleton-only baselines \citep{aganian2023object}, while temporal hand trajectories can improve assembly verification over frame-wise analysis \citep{zhou2024assembly}. Building on these findings, our system integrates region-aware interaction detection with motion trajectory verification through a unified finite-state model.

Object detection plays a vital role in assembly monitoring, though industrial environments present challenges such as small objects, reflections, occlusions, and appearance variation \citep{kupcsik2021supervisedtrainingdenseobject}. 
Modern detectors, including YOLO \citep{redmon2016lookonceunifiedrealtime, yolov8_ultralytics} and transformer-based models \citep{carion2020endtoendobjectdetectiontransformers}, involve trade-offs among accuracy and computation. In this work, RF-DETR \citep{robinson2026rfdetrneuralarchitecturesearch} performs object detection. 

The correct orientation of parts is equally crucial. 
Traditional six degrees of freedom (6-DoF) pose estimation methods \citep{wen2024foundationposeunified6dpose, florence2018denseobjectnetslearning} offer high precision but introduce significant computational complexity and occlusion sensitivity \citep{marullo20236d}. In contrast, by exploiting the physical constraints of our assembly environment, our approach bounds candidate states and reformulates general pose estimation into a lightweight, discrete task pose classification problem. In the proposed system, YOLOv8n-cls classifiers \citep{yolov8_ultralytics} are used only for task-specific pose classification, on crops produced by the detector.

Verification at the assembly stage involves detecting the correct sequence of steps. 
Current methods utilize deep learning algorithms based on temporal data \citep{zhang2025stage}, Hidden Markov Models \citep{BERG2018205}, Dynamic Time Warping (DTW) \citep{kung2022assemblydtw, TANG201821}, and finite-state machines (FSMs) for modeling procedures \citep{zhao2021fsmassembly, zhang2025stage}. While learning-based techniques demonstrate high recognition precision, they require extensive annotated datasets and can generate logically inconsistent stage transitions \citep{zhang2025stage}. Conversely, procedural models incorporate domain knowledge to guarantee execution sequence consistency \citep{zhao2021fsmassembly, BERG2018205}. 
Our framework combines learned visual evidence with finite-state procedural constraints for stage verification.

Finally, it is not obvious that more guidance is always better. The motor-learning guidance hypothesis proposes that frequent augmented feedback can improve performance during practice but may impair retention or transfer by reducing learners' processing of task-relevant information \citep{salmoni1984knowledge}. We use this hypothesis to motivate comparison of assistance levels over repeated trials; retention itself was not tested here.

%%=============================================================%%
%% 3. PROPOSED SYSTEM                                          %%
%%=============================================================%%
\section{Proposed System}\label{sec:methodology}

\subsection{System Overview}\label{subsec:system-overview}

\begin{figure}[htbp]
    \centering
\includegraphics[width=1.0\linewidth]{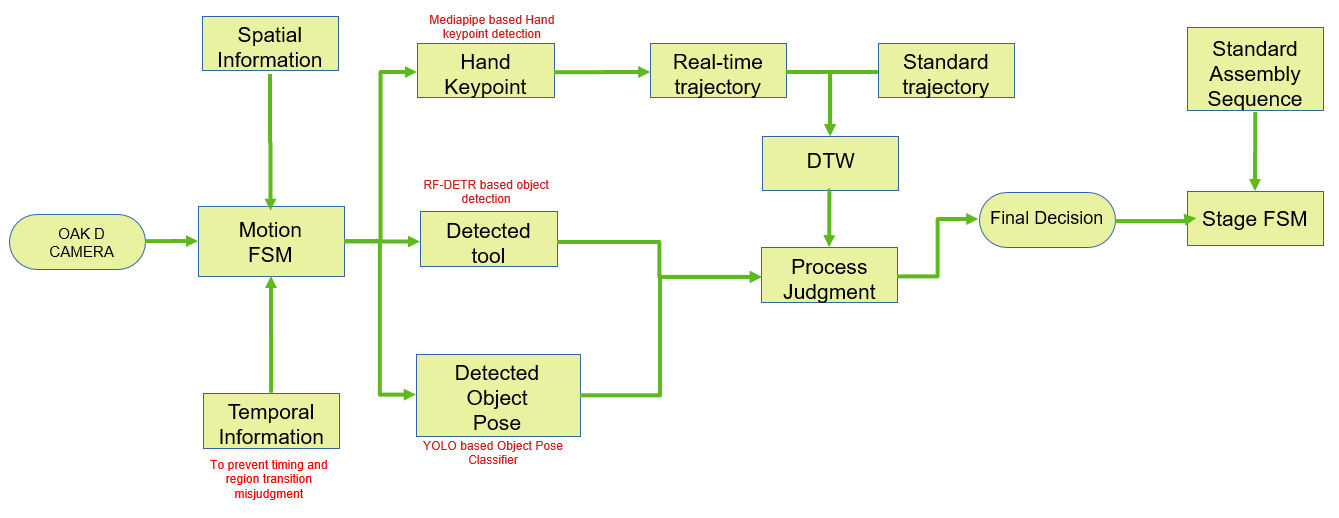}
    \caption{Overall architecture of the proposed vision-based assembly monitoring and guidance system}
    \label{fig:system-architecture}
\end{figure}

The system processes a single overhead RGB stream and separates visual perception from procedural reasoning and presentation (Figure~\ref{fig:system-architecture}). This separation defines module interfaces; it does not imply that components can be exchanged without reconfiguration or validation.

The perception layer estimates hand landmarks with MediaPipe, detects task components, classifies the orientation of selected parts, and measures image motion within configured regions of interest. These outputs are visual evidence, not assembly-state decisions. The interaction layer combines regional motion with hand-position evidence using the gated rule in Section~\ref{subsec:hybrid-roi-trigger}; temporal confirmation and cooldown reduce transitions caused by transient detections. Because these thresholds and gates trade responsiveness against spurious or missed events, the trigger must be validated for each deployment configuration.

Procedural reasoning is implemented by two finite-state machines. The MotionFSM represents the phases of an individual manipulation, while the StageFSM checks stage-completion conditions and enforces the configured task sequence. It consumes perception and interaction outputs and assigns stage outcomes; when an event is missed, the implementation permits a logged manual advancement. The guidance layer maps the verified stage and the selected assistance condition to interface content. The three conditions vary the amount of visual instruction and telemetry; the study did not implement or evaluate automatic selection of guidance based on inferred operator expertise.

\subsection{Assembly Task Representation}\label{subsec:workspace-task-rep}

The assembly procedure is modeled as a sequence of discrete stages defined in configuration files, enabling adaptation to different tasks (e.g., assembly and disassembly) without code modifications.
Each stage specifies a target component or tool, a source or operational region, expected hand motion, completion conditions, and optional duration. Completion is inferred from the next stage's pick event (\texttt{next\_stage\_pick}), an explicit placement event (\texttt{completion\_proxy}), or the terminal event (\texttt{terminal\_end}), which marks task completion.
Stages are grouped by workflow constraints. Ordered groups enforce a fixed sequence; 
unordered groups allow subtasks that permit arbitrary execution permutations, such as installing symmetric tie rods, in any valid order.

\subsection{Workspace and Region Modelling}\label{subsec:workspace-modelling}

The workspace is partitioned into a set of regions of interest (ROIs) representing physically meaningful workbench locations, stored in a configuration file. Tool and component box ROIs are defined in pixel coordinates, whereas the main workspace plane is projected into normalized metric coordinates via ArUco-marker homography. This perspective correction ensures spatial invariance, which subsequently standardizes hand trajectories for reliable motion comparison via Dynamic Time Warping (DTW).

\begin{figure}[tp]
  \centering
\includegraphics[width=0.72\linewidth]{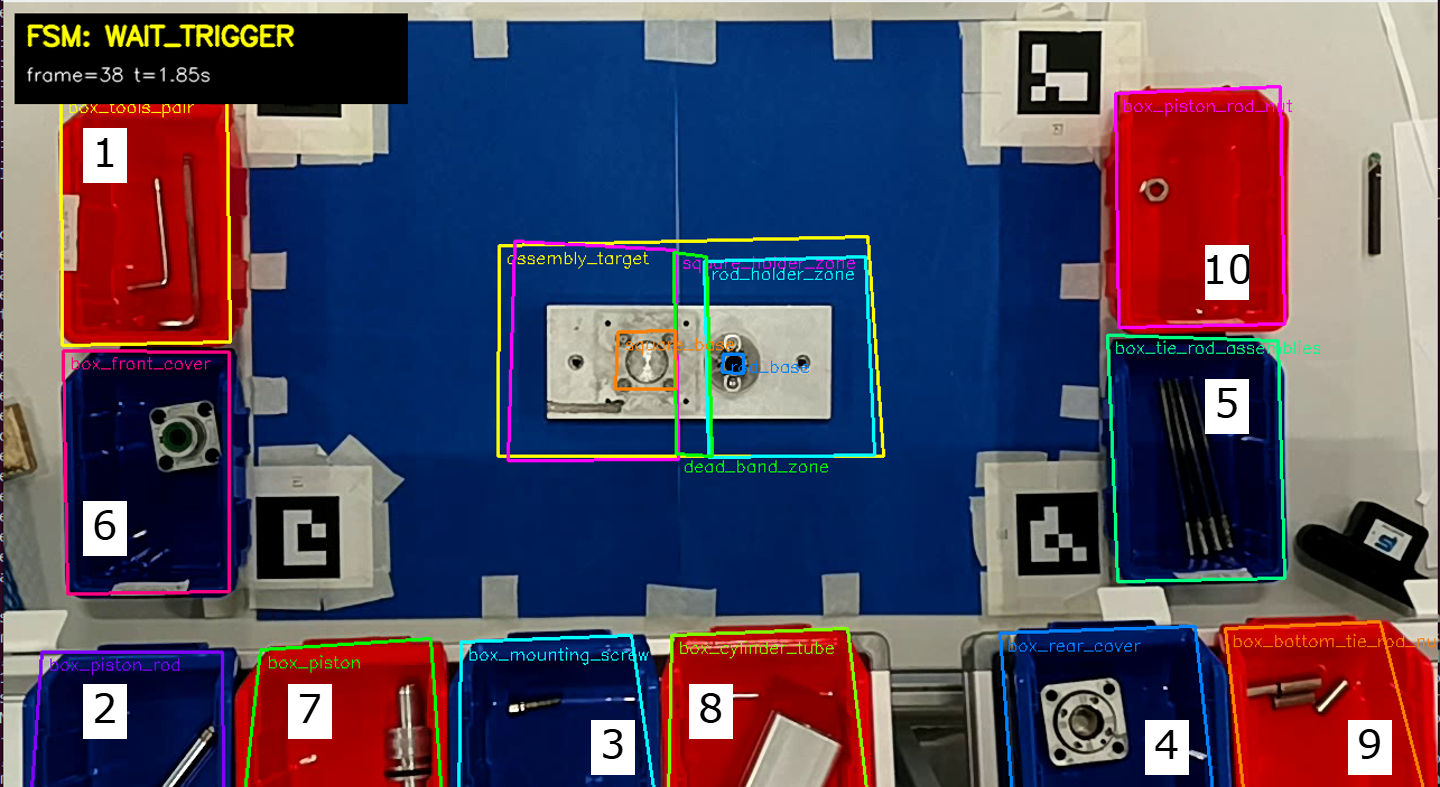}
\par\smallskip
{\small
\begin{tabular}{@{}ll@{\hspace{1.5em}}ll@{}}
\colorbox{toolyellow}{\rule{1.2ex}{1.2ex}} & [1] Tool box & \colorbox{frontpink}{\rule{1.2ex}{1.2ex}} & [6] Front cover \\
\colorbox{pistonviolet}{\rule{1.2ex}{1.2ex}} & [2] Piston rod & \colorbox{pistonlime}{\rule{1.2ex}{1.2ex}} & [7] Piston \\
\colorbox{screwcyan}{\rule{1.2ex}{1.2ex}} & [3] Mounting screw & \colorbox{tubegreen}{\rule{1.2ex}{1.2ex}} & [8] Cylinder tube \\
\colorbox{rearblue}{\rule{1.2ex}{1.2ex}} & [4] Rear cover & \colorbox{bottomorange}{\rule{1.2ex}{1.2ex}} & [9] Bottom tie-rod nut \\
\colorbox{assembliesgreen}{\rule{1.2ex}{1.2ex}} & [5] Tie-rod nut assemblies & \colorbox{rodnutpink}{\rule{1.2ex}{1.2ex}} & [10] Nut
\end{tabular}}
\caption{Workspace regions of interest (ROIs). Color swatches identify the regions shown in the image}
  \label{fig:regions}
\end{figure}

These regions serve three core purposes: providing spatial interaction constraints, binding assembly stages to specific zones, and supporting the guidance overlay. A central dead-band zone filters transient hand movements between active regions.

\subsection{Hybrid ROI Interaction Detection}\label{subsec:hybrid-roi-trigger}

The system uses two distinct cues for each region: image motion and hand presence. Motion alone can be caused by global arm movement, while hand position alone can remain active during brief tracking artifacts. The implemented trigger therefore requires both motion in the region and confirmation by either of two hand-position gates:
\begin{equation}
\text{Trigger}(R_i) = M_i \land H_i, \qquad H_i = G_{1,i} \lor G_{2,i}
\label{eq:trigger_logic}
\end{equation}
Here $M_i$ is motion evidence in region $R_i$, while $G_{1,i}$ and $G_{2,i}$ are the consecutive-frame and rolling-window hand-position gates defined below. Once the conjunction is satisfied, temporal confirmation and cooldown are applied.

To prevent flickering from noisy frame outputs, validated interactions must persist unbroken through a confirmation window $\tau_{\text{confirm}}$ before triggering state transitions, followed by a cooldown period $\tau_{\text{cooldown}}$:
\begin{equation}
\text{Trigger}(R_i) = (M_i \land H_i) \xrightarrow{\text{hold } \tau_{\text{confirm}}} \text{Fire} \xrightarrow{\text{hold } \tau_{\text{cooldown}}} \text{Reset}.
\label{eq:temporal_confirmation}
\end{equation}
The output signal feeds directly into the MotionFSM to govern manipulation start and completion boundaries, decoupling low-level spatial activity detection from procedural stage logic.

\subsubsection{Motion Evidence}\label{subsubsec:visual-activity}

Motion is extracted by background subtraction relative to the setup baseline $B(x,y) = \text{median}\{I_1(x,y), \dots, I_N(x,y)\}$. For frame $I_t(x,y)$, the binary motion map $D(x,y)$ is defined using pixel threshold $\theta_{\mathrm{pix}}$:
\begin{equation}
D(x,y) = \mathbb{I}[|I_t(x,y) - B(x,y)| > \theta_{\mathrm{pix}}].
\label{eq:binary_motion}
\end{equation}
The raw motion score $m_i$ for region $R_i$ is the fraction of active pixels, $m_i = \frac{1}{|R_i|} \sum_{(x,y) \in R_i} D(x,y)$. To reduce false triggers from global arm sweeps, the effective score $\tilde{m}_i$ applies global suppression with a strong-motion exemption:
\begin{equation}
\tilde{m}_i =
\begin{cases}
m_i, & \text{if } m_{\text{global}} \le T_{\text{supp}} \text{ or } m_i > T_{\text{strong}} \\
\alpha \cdot m_i, & \text{otherwise}
\end{cases}
\label{eq:motion_suppression}
\end{equation}
where $m_{\text{global}}$ is the frame-wide motion fraction, $T_{\text{supp}}$ is the suppression threshold, $\alpha \in (0, 1)$ is the discounting factor, and $T_{\text{strong}}$ preserves localized pick activity during background motion. Motion evidence is $M_i = \mathbb{I}[\tilde{m}_i > \theta_m]$, where $\theta_m=0.01$ is the minimum fraction of changed pixels in a region.

\subsubsection{Hand-Position Evidence}\label{subsubsec:hand-evidence}

Hand presence is tracked using an anchor point $P_{\text{hand}} = \frac{1}{2}(P_{\text{wrist}} + P_{\text{index\_mcp}})$, providing spatial stability during partial occlusions or tool handling. Region interaction is validated via two parallel gates:
\begin{equation}
G_{1,i}: P_{\text{hand}} \in R_i \text{ for } \ge k \text{ consecutive frames},
\label{eq:hand_gate}
\end{equation}
\begin{equation}
G_{2,i}: \sum_{t \in W} \mathbb{I}[P_{\text{hand}}(t) \in R_i] \ge n \text{ within rolling window } W.
\label{eq:traj_gate}
\end{equation}
The first gate filters brief transit crossings by requiring consecutive presence; the second tolerates temporary MediaPipe dropouts within window $W$. 
Together they define $H_i = G_{1,i} \lor G_{2,i}$, the hand-position evidence in Eq.~\eqref{eq:trigger_logic}.

\subsection{Motion Finite-State Machine}\label{subsec:motion-fsm}

\begin{figure}[htbp]
    \centering
\includegraphics[width=0.72\linewidth]{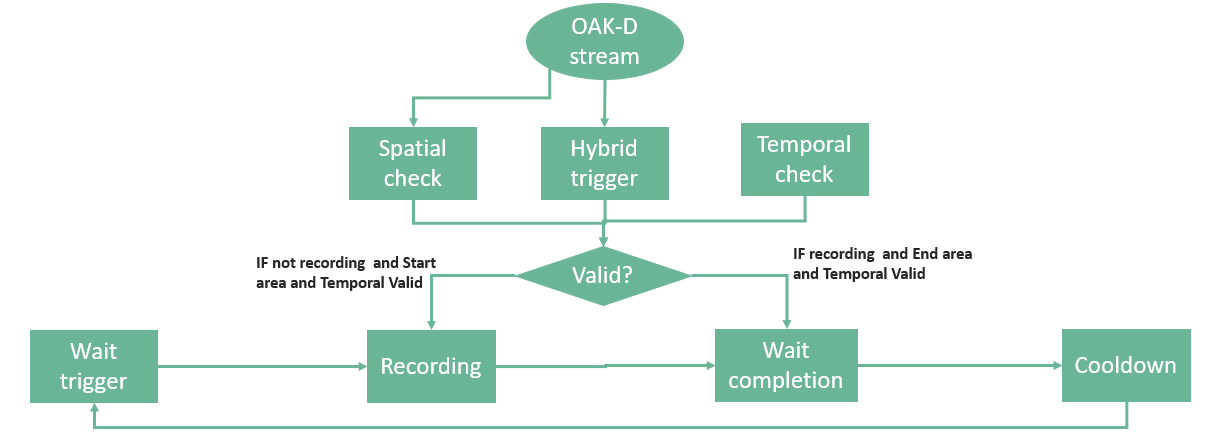}
\caption{Architecture of the motion finite-state machine}
    \label{fig:motionfsm}
\end{figure}

The MotionFSM models each manipulation as a progression through four states (Figure~\ref{fig:motionfsm}): \texttt{\seqsplit{WAIT\_TRIGGER}}, \texttt{\seqsplit{RECORDING}}, \texttt{\seqsplit{WAIT\_COMPLETION\_CONFIRM}}, and \texttt{\seqsplit{COOLDOWN}}.

\textbf{\texttt{\seqsplit{WAIT\_TRIGGER.}}} The FSM waits for the trigger defined in Section~\ref{subsec:hybrid-roi-trigger}, the expected stage-specific start role (for example, \texttt{\seqsplit{part\_pick}} or \texttt{\seqsplit{tool\_pick}}) in the source region, and confirmation that the interaction occurs in the expected workspace region.

\textbf{\texttt{\seqsplit{RECORDING.}}} On entry, active-hand trajectory recording begins and the stage completion mode defined in Section~\ref{subsec:workspace-task-rep} is evaluated. 
Completion may be inferred from the next pick (\texttt{\seqsplit{next\_stage\_pick}}), an explicit region event (\texttt{\seqsplit{completion\_proxy}}), or the task's terminal condition (\texttt{\seqsplit{terminal\_end}}). 
If the terminal event is missed or between stage completion trigger fails, the operator may manually advance the FSM and the intervention is logged. Brief hand-tracking dropouts are tolerated for up to 10 frames.

\textbf{\texttt{\seqsplit{WAIT\_COMPLETION\_CONFIRM.}}} Triggered upon completion candidate detection, this state enforces signal persistence over a short frame window to filter single-frame vision noise. Any signal drop resets the state to \texttt{\seqsplit{RECORDING}}, preventing premature stage closure from transient flickers.

\textbf{\texttt{\seqsplit{COOLDOWN.}}} Following confirmed completion, the FSM holds for a fixed frame duration to let physical motion settle and prevent residual actions from falsely triggering subsequent stages, completing a two-stage temporal smoothing pipeline.

\subsection{Stage Finite-State Machine}\label{subsec:stage-fsm}

\begin{figure}[htbp]
    \centering
\includegraphics[width=0.72\linewidth]{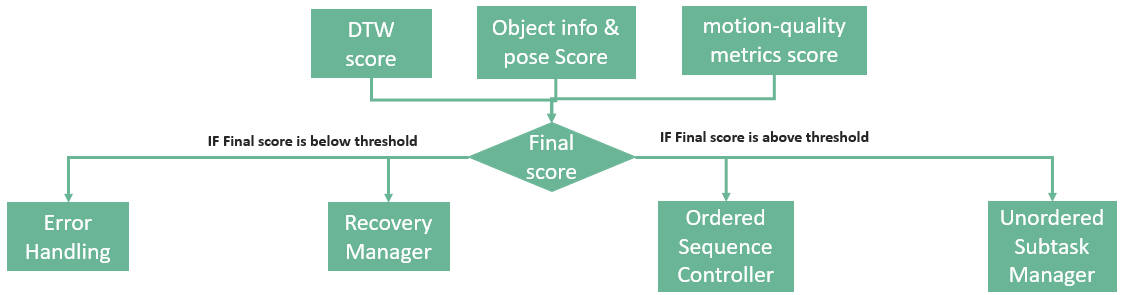}
\caption{Architecture of the stage finite-state machine}
    \label{fig:stagefsm}
\end{figure}

While the MotionFSM handles low-level manipulations, the StageFSM governs high-level procedural logic.
As shown in Figure~\ref{fig:stagefsm}, input evidence comprising DTW trajectory alignment (Section~\ref{subsec:trajectory-verification}), object info and pose scores, and motion-quality metrics is aggregated into a composite score.
Depending on whether this score satisfies the decision threshold, execution routes to failure-handling or procedural advancement branches.

\subsubsection{Stage Transitions and Failure Handling}

Stages transition through \texttt{PENDING}, \texttt{ACTIVE}, and \texttt{PASSED} or \texttt{FAILED} states. A stage transitions to \texttt{PASSED} only when all multi-modal verification conditions (region bounds, pose/orientation, DTW, tool presence) hold simultaneously. If the final score falls below the threshold, execution routes to downstream error components:
The error handler assigns a reason code (for example, missing or incorrect object, tool, orientation, trajectory, or sequence) and displays corresponding feedback. Recovery permits another attempt, manual FSM advancement after a missed event (with the intervention logged), or a reset of the current stage or full task.

\subsubsection{Sequence Control and Unordered Management}

After stage validation, sequence control checks the task definition. Ordered groups advance step by step and flag out-of-order interactions. Unordered groups accept any remaining subtask that meets its completion conditions and continue until the group is complete.

\subsection{Hand-Trajectory Verification}\label{subsec:trajectory-verification}

Trajectory verification compares the observed motion to an expert reference and extracts quantitative motion-quality metrics.

MediaPipe hand landmarks are processed once per incoming camera frame. 
The default OAK-D capture rate is 20 frames per second (FPS), therefore MediaPipe trajectories are sampled at approximately 20 Hz. Invalid points are removed, single-frame spikes suppressed, and missing samples interpolated. A Savitzky--Golay filter with a 5-sample window and polynomial order 2 smooths each coordinate. When both hands are visible, the active hand is selected using dominant-hand preference, velocity, and tool proximity, with hysteresis to reduce contamination from the inactive hand.

\begin{figure}[htbp]
    \centering
\includegraphics[width=0.43\linewidth]{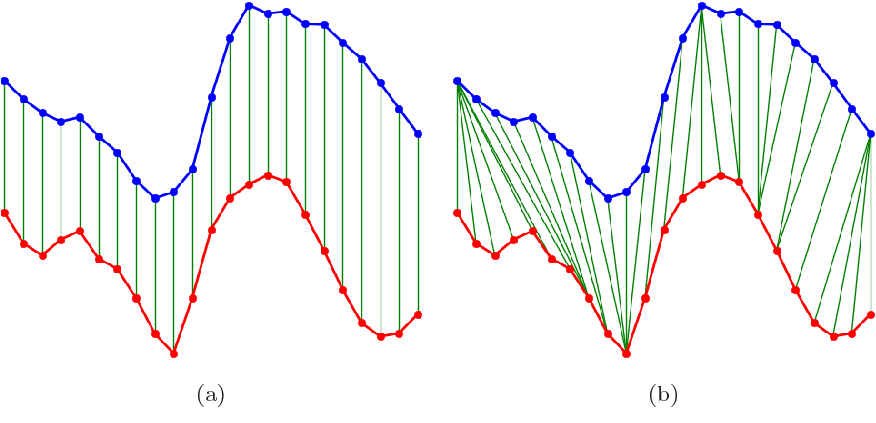}
    \caption{Comparison of (a) Euclidean matching vs.\ (b) dynamic time warping matching}
    \label{fig:dtw}
\end{figure}

The recorded trajectory is compared with an expert reference using Dynamic Time Warping (DTW). DTW permits local temporal stretching, so motions with different durations can still be aligned (Figure~\ref{fig:dtw}). Formally, for observed trajectory $P$ and reference $Q$, the distance is defined recursively as
\begin{equation}
    DTW(i,j) = d(i,j) + \min
    \begin{cases}
        DTW(i-1,j),\\
        DTW(i,j-1),\\
        DTW(i-1,j-1),
    \end{cases}
    \label{eq:dtw}
\end{equation}
where $d(i,j)$ is the Euclidean distance between samples $p_i$ and $q_j$, and the final DTW distance is $DTW(n,m)$ for trajectories of lengths $n$ and $m$. This distance serves a dual role: as a stage gate, marking trajectories too dissimilar from the reference as \texttt{FAILED} and prompting the operator to repeat the stage; and as a standalone metric for the experimental evaluation, providing a continuous quality measure independent of pass/fail outcomes.

Four lower-is-better metrics evaluate active-hand motion quality: jerk index is the mean absolute finite-difference jerk, $\text{mean}(|j_i|)$, with normalized-coordinate units s$^{-3}$; pause ratio is the fraction of stage time with speed below 0.02 normalized-coordinate units s$^{-1}$; micro-stop count is the number of brief velocity dips lasting 0.05--0.25 s; and path inefficiency is the ratio of observed path length to straight-line displacement. These metrics support offline system evaluation and user-study analysis.

\subsection{Object Detection}\label{subsec:object-detection}

Object detection provides the spatial evidence needed to identify which component or tool is being manipulated.

\subsubsection{Object Classes}\label{subsubsec:object-classes}

\begin{figure}[htbp]
  \centering
\includegraphics[width=0.68\linewidth]{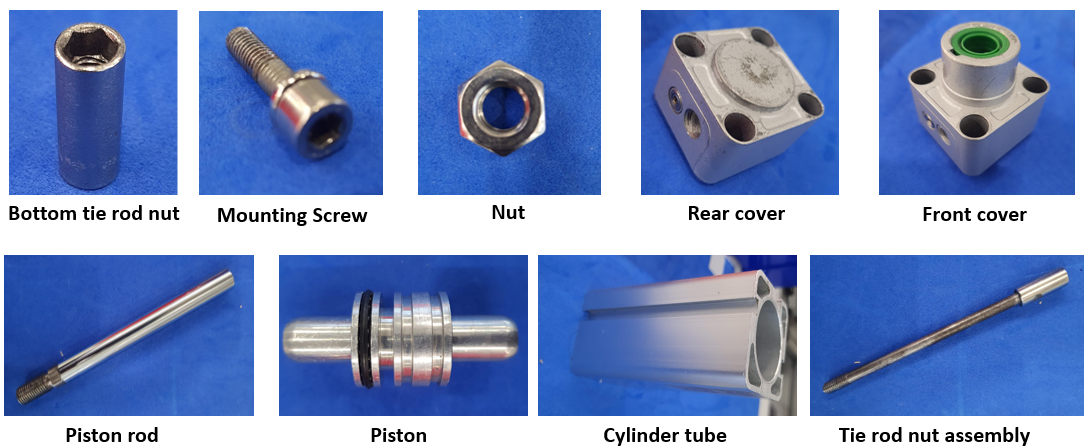}
  \caption{Target object classes and assembly components}
  \label{fig:object_classes}
\end{figure}
Figure~\ref{fig:object_classes} shows the nine components, and additional two screwdriver classes were used by the detector.
\begin{table}[htbp]
\centering
\caption{The 11 detector classes used in training.}
\label{tab:object-classes}
\small
\begin{tabular}{ll}
\toprule
Component classes & Tool classes \\
\midrule
Bottom tie-rod nut & Small L-shaped screwdriver \\
Mounting screw & Large L-shaped screwdriver \\
Nut  & \\
Rear cover & \\
Front cover & \\
Piston rod & \\
Piston & \\
Cylinder tube & \\
Tie-rod nut assembly & \\
\bottomrule
\end{tabular}
\end{table}
We employ a task-specific class set to maximize detection performance rather than relying on generic object detectors. 
Preliminary evaluations with off-the-shelf models indicated insufficient reliability when distinguishing visually similar components such as front and rear covers, which share comparable size, color, and 
geometry particularly under partial occlusions by the operator's hand.

\subsubsection{Training Dataset and Inference}\label{subsubsec:object-detection-dataset}

\begin{figure}[htbp]
  \centering
\includegraphics[width=0.68\linewidth]{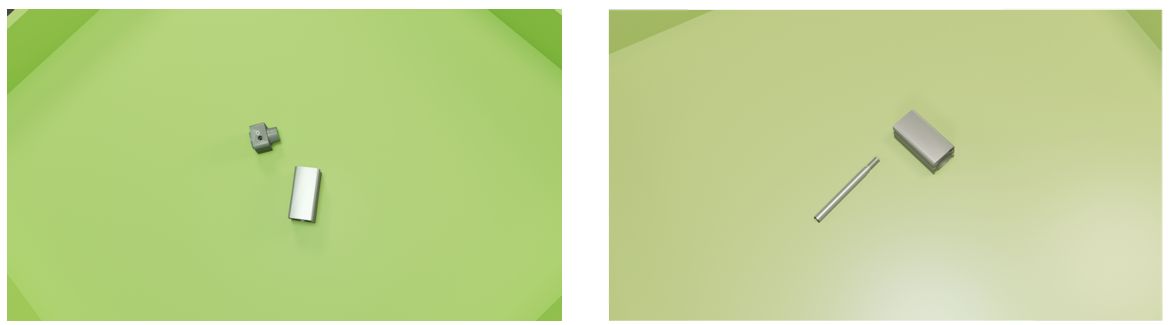}
  \caption{Examples of CAD-rendered images generated for synthetic-data augmentation experiments}
  \label{fig:synthetic_training_data}
\end{figure}

The deployed RF-DETR Small detector was trained on the real-image COCO dataset (\texttt{\seqsplit{Assembly\_parts\_detection.v2i.coco}}), collected under the overhead camera and split into 942 training, 241 validation, and 324 test images. Training used RF-DETR~1.3.0 with an input resolution of 512 pixels for 120 epochs, a batch size of 4 with gradient accumulation over 2 steps (effective batch size 8), and a learning rate of $10^{-4}$.
Synthetic images were prepared separately for data-augmentation experiments (Figure~\ref{fig:synthetic_training_data}). They were rendered from Computer-Aided Design (CAD) models by varying object pose, scale, lighting, and background, and require no manual annotation; the mixed-data experiments added 123 or 184 synthetic images to the real training set. The deployed checkpoint does not come from these experiments.

YOLO (v8, v11, v26), RTMDet, and RF-DETR were compared during detector selection. RF-DETR Small is the object detector in the deployed system. Detection inference is offloaded to a remote server.

\subsection{Task-Specific Pose Classification}\label{subsec:pose-classification}

\begin{figure}[htbp]
    \centering
\includegraphics[width=1.00\linewidth]{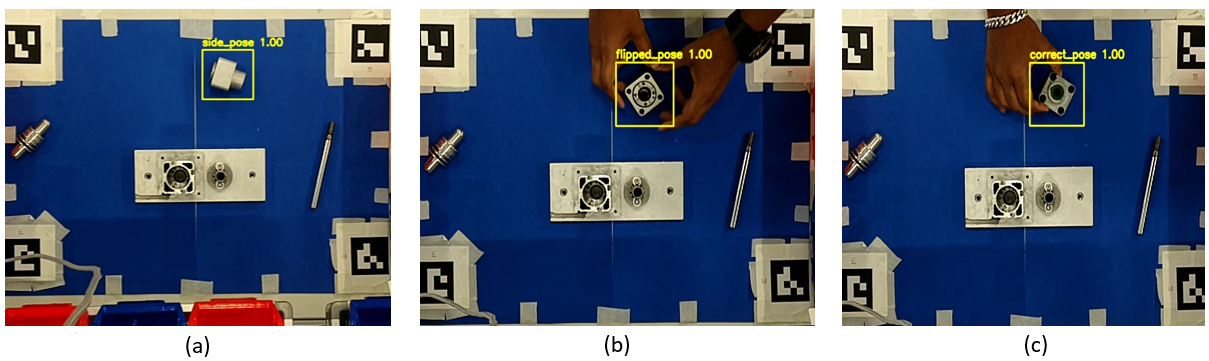}
\caption{YOLOv8n-cls pose classification for the front cover: (a) side pose, (b) flipped pose, and (c) correct pose}
    \label{fig:Yolo_Pose_Classifer}
\end{figure}

Component orientation is critical for several assembly stages, such as placing cylinder covers face-up or inserting the piston rod correctly. 
Because standard bounding boxes convey no rotational state, an orientation verification step is required before marking a stage passed. 
As reflective metal surfaces produced depth dropouts and sparse point clouds, depth-based pose estimation was not used. 
In a preliminary FoundationPose test across 4 synthetic object crops evaluated under clean, mild, and heavy depth corruption protocols, pose estimation failed when valid depth coverage within the object mask fell below about 71\% (dropping to a median valid depth of 65.7\% under heavy corruption, which triggered pose initialization and CAD matching failures).
Rotational symmetry also led to invalid 180-degree flips. 
We instead classify task-relevant orientations (Figure~\ref{fig:Yolo_Pose_Classifer}). 

\subsubsection{Pose-Class Definition}\label{subsubsec:pose-class-definition}

For components requiring orientation verification, a small set of valid assembly orientations is defined based on overhead appearance, while axially or geometrically symmetric parts (screws, nuts, tie-rods) are excluded, because their physical geometry constrains them to a single valid orientation. The discrete classes comprise:
\textbf{Front Cover} (\textit{correct} [sealing face up], \textit{sideway}, \textit{flipped} [sealing face down]);
\textbf{Rear Cover} (\textit{correct} [sealing face up], \textit{sideway}, \textit{flipped} [sealing face down]); and
\textbf{Piston Rod} (\textit{correct} [forward insertion], \textit{sideway}, \textit{reversed}).

\subsubsection{Training Dataset and Classifier Architecture}\label{subsubsec:pose-training-dataset}

Classifiers were trained on real image crops produced by the object detector (the pose classifier dataset), with standard augmentation.
Each classifier is a YOLOv8n-cls model trained for 60 epochs at an input size of $224\times224$ pixels with a batch size of 32; the deployed models use the selected \texttt{best.pt} checkpoints, which record Ultralytics~8.4.51.
A separate per-component classifier predicts one of the defined pose classes. The classifier is invoked only for components requiring orientation verification, and only when the detector produces a confident detection. 
The StageFSM applies the classification result at stage completion.
The piston-rod classifier's validation performance was substantially lower than that of the front and rear cover classifiers, and is therefore treated as a
preliminary component. It was nevertheless included in the
experimental system, and its predictions were available to
the StageFSM during piston-rod-related stage verification.
Its performance is reported separately and is not considered
quantitatively validated in this study.

\subsection{Adaptive Guidance Interface}\label{subsec:adaptive-interface}

The guidance interface consumes the outputs of the perception modules and both FSMs and renders a live visual overlay that adapts to the current stage and the selected assistance level, without affecting perception or verification.
Three assistance levels are defined.

\subsubsection{Full guidance condition}\label{subsubsec:full-assistance}

\begin{figure}[htbp]
    \centering
\includegraphics[width=0.75\linewidth]{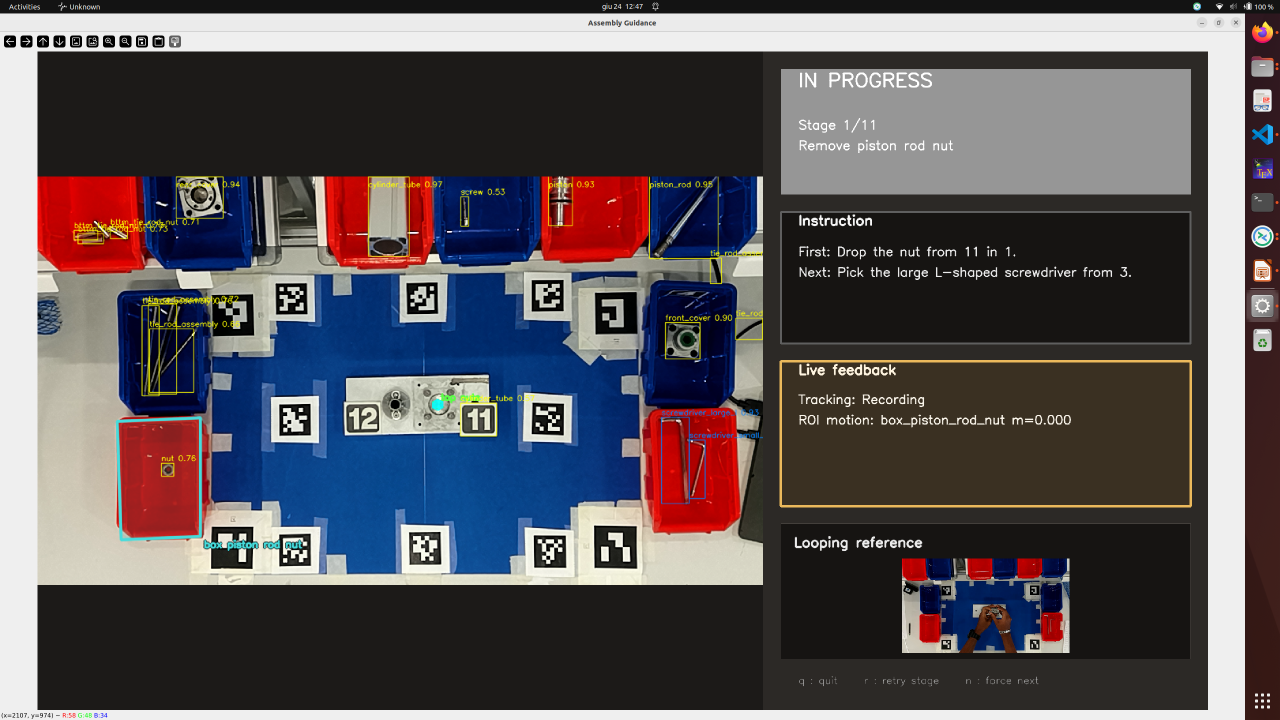}
\caption{Full guidance mode}
    \label{fig:fassistance}
\end{figure}
Designed for novice operators, full-guidance mode (Figure~\ref{fig:fassistance}) provides visual feedback: a light-blue bounding box marks the target part-pick ROI, while a cyan CAD projection indicates the placement pose. A grey status banner monitors stage progression (green for success, red for error), alongside a yellow telemetry panel that logs motion activity, orientation, and FSM states. Step-by-step text instructions and a looping expert reference video provide additional guidance.

\subsubsection{Mid guidance condition}\label{subsubsec:medium-assistance}

\begin{figure}[htbp]
    \centering
\includegraphics[width=0.75\linewidth]{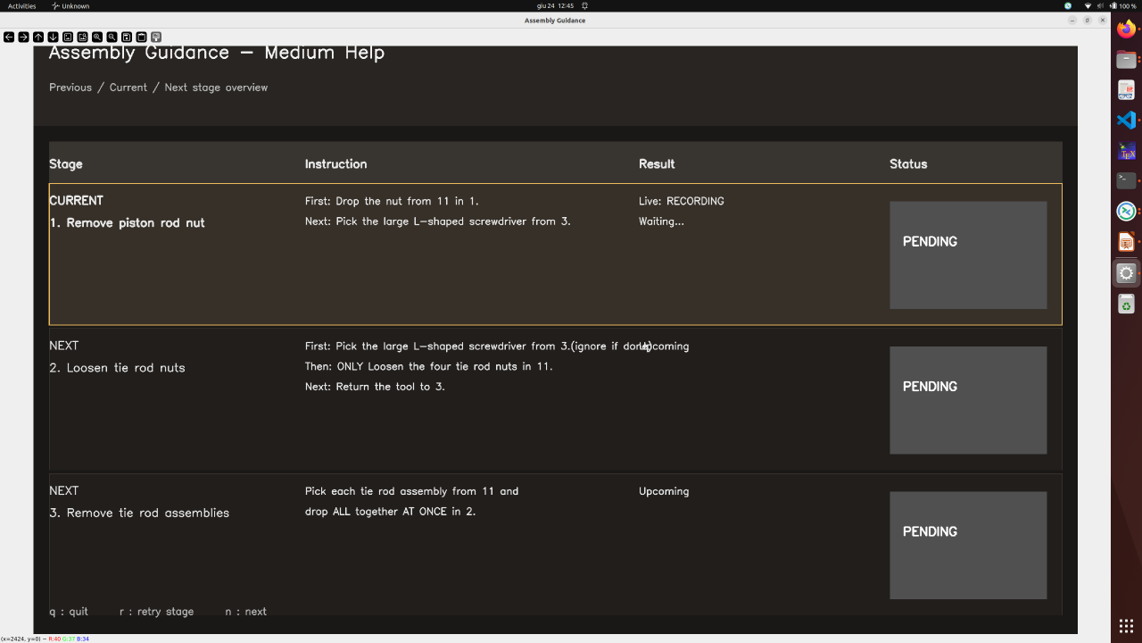}
\caption{Mid guidance mode}
    \label{fig:massistance}
\end{figure}
Mid-guidance mode (Figure~\ref{fig:massistance}) presents selective feedback. A highlighted yellow row tracks the active stage in the guidance table. The \textit{Instructions} column describes the steps, the \textit{Results} column logs telemetry (orientation, motion activity, FSM state), and the \textit{Status} column records stage pass/fail outcomes.

\subsubsection{No guidance condition}\label{subsubsec:no-assistance}

\begin{figure}[htbp]
    \centering
\includegraphics[width=0.75\linewidth]{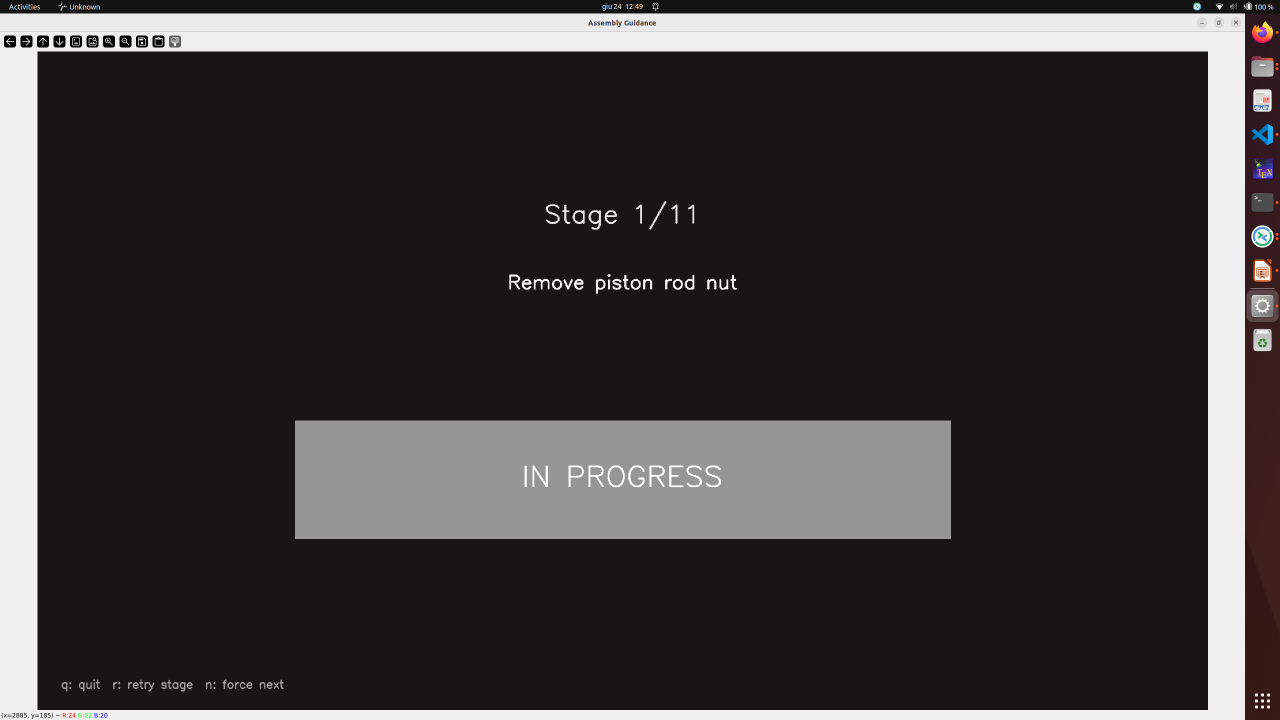}
\caption{No-guidance mode}
    \label{fig:nassistance}
\end{figure}
No-guidance mode (Figure~\ref{fig:nassistance}) establishes an unassisted baseline by hiding continuous guidance and live telemetry while keeping perception modules active in the background. The interface displays only the active stage index, descriptor name, and a central status tab indicating state progression (grey for ongoing, green for success, red for re-execution).

%%=============================================================%%
%% 4. EXPERIMENTAL SETUP                                       %%
%%=============================================================%%
\section{Experimental Setup}\label{sec:experimental-setup}

\subsection{Hardware and Software Configuration}\label{subsec:hardware-software}

\begin{figure}[htbp]
  \centering
\includegraphics[width=0.52\linewidth]{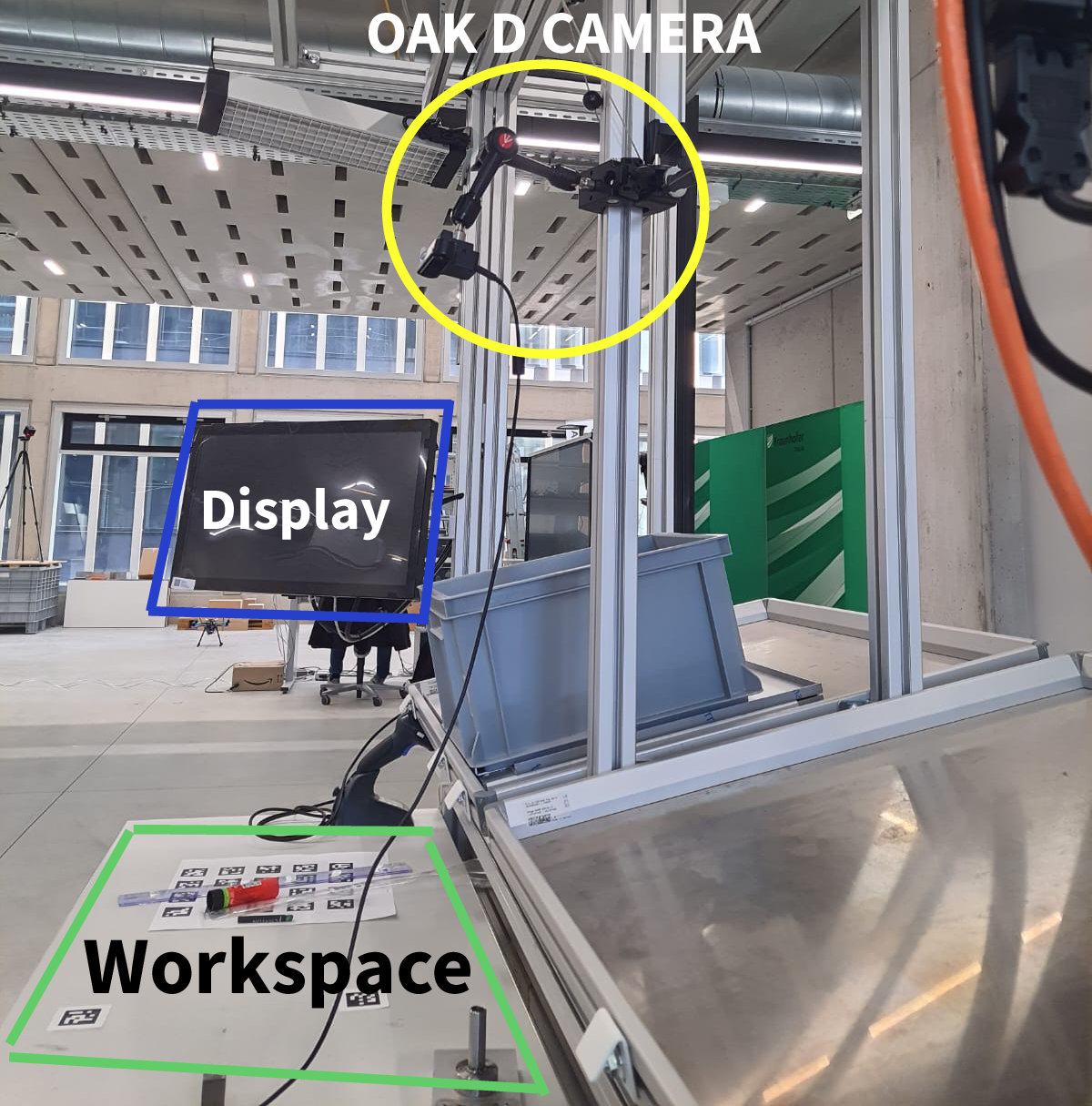}
  \caption{Laboratory workstation setup}
  \label{fig:hardware_setup}
\end{figure}

The system runs on a fixed workstation in a laboratory environment at Fraunhofer Italia. A single overhead red-green-blue plus depth (RGB-D) camera (Luxonis OAK-D) is mounted directly above the assembly workspace, providing a top-down view of the workbench and the operator's hands, as shown in Figure~\ref{fig:hardware_setup}.
Depth was not used as an input to the deployed perception pipeline because reflective surfaces and occlusions produced unreliable depth maps; the system therefore uses the camera's RGB stream. Deep-learning inference, detection, and pose classification run on a remote server equipped with an RTX 3090 graphics processing unit (GPU), connected over a local network.
The stack uses Python 3.11, OpenCV, MediaPipe, PyTorch, Ultralytics, and the RF-DETR reference implementation. The FSMs, trajectory processing, and guidance rendering are implemented in custom Python modules.

\subsection{Assembly Task and Dataset}\label{subsec:experimental-task}

\begin{figure}[htbp]
  \centering
\includegraphics[width=0.58\linewidth]{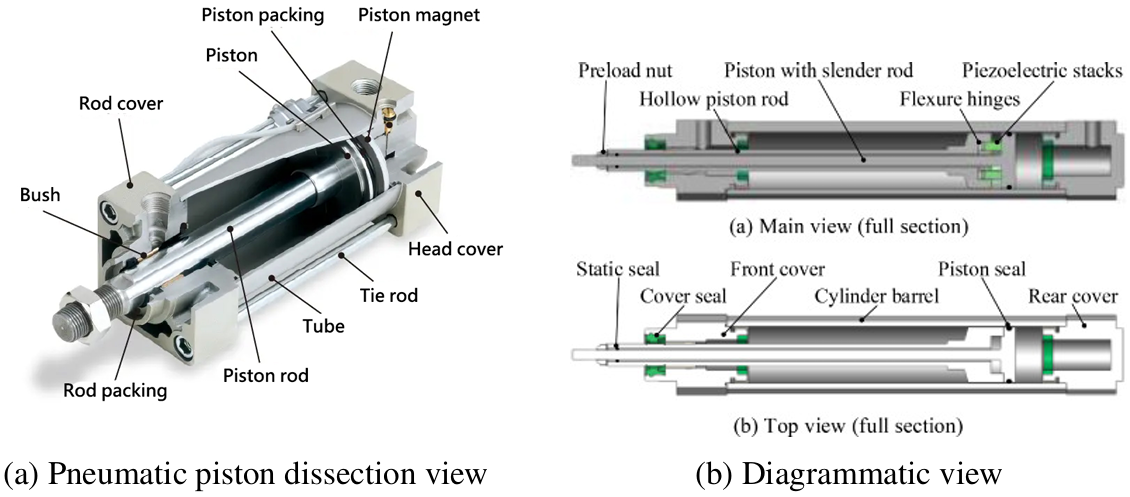}
  \caption{The pneumatic piston used in the evaluation provided by University of Bolzano}
  \label{fig:pneumatic_piston_dissection}
\end{figure}

The evaluation task was assembly and disassembly of the pneumatic piston in Figure~\ref{fig:pneumatic_piston_dissection}. The assembly workflow comprised 11 stages, whereas the disassembly
workflow comprised 10 stages. Each workflow included ordered stages
and unordered groups, as defined by the task configuration. 
Evaluation used the validation split of the detector dataset, the validation crops of the pose classifier dataset, and numerical reference trajectories for DTW. One expert reference trajectory was used per elementary action, yielding 11 assembly and 11 disassembly references; the mounting-screw and piston drop actions have separate references, D08a and D08b.

\subsection{Evaluation of Technical Components}\label{subsec:technical-evaluation}

Object detection was evaluated on the validation split of the real-image detector dataset (241 images) using precision, recall, and mean average precision at an intersection-over-union threshold of 0.5 (mAP@0.5).
Pose classifiers were evaluated by top-1 accuracy on the validation crops of the pose classifier dataset, as recorded in the training logs; for the front and rear covers, the reported values are final-epoch validation accuracies.
For stage-trigger review, each false-negative code in the manual intervention logs was counted as a missed event, with the analyzed stage records in that condition as the denominator (207 full, 227 mid, 243 no guidance). The logs were not an independently annotated ground-truth set, so true-positive and false-trigger rates are not reported.

\subsection{User Study and Protocol}\label{subsec:user-study}

Eight uncompensated, voluntary participants (employees of Fraunhofer Italia Research, with no prior assembly experience) completed the study. Session metadata were anonymized using numeric identifiers. 
Three assistance levels (Full, Mid, No) were evaluated, corresponding to the guidance interfaces in Section~\ref{subsec:adaptive-interface}. The level was assigned per trial in a pseudo-randomized order, with the constraint that each participant experienced all three levels within their first three trials.

Participants completed 6--12 trials. The parsed study records do not document a prespecified or self-reported stopping criterion, so the reason sessions ended cannot be determined from these data. Conditions alternated in a pseudo-randomized within-participant order, allowing skill acquired in one condition to carry over to later conditions. 
Thus, each condition-specific curve describes performance under that condition over total session experience, not learning isolated to that condition. 
Participants performed assembly and disassembly trials ending at the terminal state; experimenter clarifications, corrections, and manual FSM advancements were recorded as interruptions. The experiment directory contained 72 participant-recording folders. Two folders with no FSM decisions were excluded, leaving 70 valid trials from eight participants; the separately conducted author demonstration was outside this participant-trial set. The 70 trials contained 698 stage decisions; 21 stage rows with one or more missing motion metrics were excluded, leaving 677 observations for motion-quality analysis. Cycle participation was 8 participants for cycles 1--6, 7 for cycles 7--8, 3 for cycles 9--10, and 1 for cycles 11--12.

\subsection{Objective Performance Measures}\label{subsec:objective-measures}

Four objective measures were extracted (Table~\ref{tab:objective_measures}). Active completion time sums the recorded stage intervals and excludes inter-stage waiting. Assembly errors comprise FSM stage failures and experimenter interventions. Active-hand trajectories provide four lower-is-better measures: jerk index, pause ratio, micro-stop count, and path inefficiency. To combine these trajectory measures $m \in \{\text{jerk}, \text{pause}, \text{stops}, \text{ineff}\}$, each stage value $x_{i,m}$ was globally standardized:
\begin{equation}
z_{i,m} = \frac{x_{i,m} - \bar{x}_m}{\sigma_m}
\end{equation}
where $\bar{x}_m$ and $\sigma_m$ are global mean and standard deviation. Second, because metrics are lower-is-better, values were sign-inverted to denote positive quality ($q_{i,m} > 0$ indicates above-average quality):
\begin{equation}
q_{i,m} = -z_{i,m}
\end{equation}
Third, inverted values were averaged row-wise over available metrics $M_i$ ($|M_i| = 4$):
\begin{equation}
\texttt{\seqsplit{skill\_score}}_i = \frac{1}{|M_i|} \sum_{m \in M_i} q_{i,m}
\end{equation}
The score is dimensionless and zero-centered, but its standard deviation is not necessarily one. Completion time and errors were analyzed separately because they are trial-level speed and correctness outcomes, whereas the composite describes stage-level movement quality. Condition contrasts are therefore reported in composite-score units.

\begin{table}[htbp]
\centering
\caption{Summary of the experimental dataset and objective measures. Failed-stage decisions count unsuccessful FSM decisions; in these data, the 29 decisions corresponded to 29 unique failed stages.}
\label{tab:objective_measures}
\small
\begin{tabular}{lcccc}
\toprule
\textbf{Metric / Category} & \textbf{Full Guidance} & \textbf{Mid Guidance} & \textbf{No Guidance} & \textbf{Total} \\
\midrule
Valid Trials ($N$) & 21 & 22 & 27 & 70 \\
Stage Observations & 207 & 227 & 243 & 677 \\
Mean Duration (s) & $237.7 \pm 138.8$ & $197.6 \pm 61.2$ & $177.7 \pm 80.8$ & --- \\
Failed-stage decisions & 11 & 9 & 9 & 29 \\
FSM False Negatives & 7 (3.4\%) & 2 (0.9\%) & 2 (0.8\%) & 11 \\
\botrule
\end{tabular}
\end{table}

\subsection{Subjective Measures}\label{subsec:subjective-measures}

Two questionnaires were administered. After each trial, participants rated four 7-point Likert items capturing perceived usefulness, ease of understanding, self-assessed independence, and reliance behavior. At the end of the session, a 6-question debrief survey evaluated preferences for hypothetical real-world deployment, distinguishing learning-time preference from daily-use preference.

\subsection{Statistical Analysis}\label{subsec:statistical-analysis}

\paragraph{Mixed-effects modeling.}
The reported stage-level models include participant and trial-within-participant random intercepts, with residual error, and were fitted by maximum likelihood (ML). The primary model is:
\[
\begin{aligned}
\mathrm{skill\_score}_i = {} & \beta_0 + \beta_1 \cdot \mathbb{I}[\mathrm{mid}] + \beta_2 \cdot \mathbb{I}[\mathrm{no}] + \beta_3 \cdot \log(\mathrm{trial}_i) + \beta_4 \cdot \mathrm{interruptions}_i \\ 
& + \beta_5 \cdot I[dismantle] + \Sigma_s \beta_s \cdot I[stage\_order\_i = s] + u_j + v_{jk} + \epsilon_i
\end{aligned}
\]
where full guidance is the reference condition, log trial number models decelerating change, and interruptions control for experimenter intervention. 
The interaction model adds condition-by-log-trial terms; a reduced model without condition indicators provides the nested baseline. 
Stage position and task type (assembly/disassembly) were included as fixed effects. Condition-specific random slopes were attempted but produced a non-positive-definite Hessian and boundary variance estimates; they were therefore not used for inference. 
The participant and trial-within-participant random intercepts were retained.

\paragraph{Variance decomposition.}
Marginal and conditional \(R^{2}\) were computed using the Nakagawa formula
and are reported with the primary model results in Section~\ref{subsec:primary-result}.
The conditional value includes both the participant and trial-within-participant random-intercept variances.

\paragraph{Likelihood-Ratio Tests and Learning Curves.} Two likelihood-ratio tests (LRTs; $\alpha = 0.05$) compare model fits using the likelihood-ratio (LR) statistic:
\begin{equation}
\text{LR} = 2(\ell_{\text{full}} - \ell_{\text{reduced}})
\label{eq:likelihood_ratio}
\end{equation}
where $\ell_{\text{full}}$ and $\ell_{\text{reduced}}$ are maximized log-likelihoods, evaluated against a $\chi^2$ distribution with degrees of freedom equal to the parameter difference. The tests evaluate: (1) \textit{Assistance effect} (primary versus reduced model), and (2) \textit{Condition $\times$ learning interaction} (interaction versus primary model). Fitted interaction parameters project learning curves across conditions with interruptions held at mean values.

\paragraph{Software.}
The statistical analyses are implemented in Python using statsmodels and SciPy; the released reproducibility environment uses Python~3.11, statsmodels~0.14.5, SciPy~1.13.1, pandas~2.2.3, and NumPy~1.26.4.

%%=============================================================%%
%% 5. RESULTS                                                  %%
%%=============================================================%%
\section{Results}\label{sec:results}

\subsection{Perception and Verification Baseline}
\label{subsec:technical-baseline}

Before the user study, perception and trigger performance were measured on the validation splits of the detector and pose classifier datasets and on reviewed study logs (Table~\ref{tab:technical-baseline}).

\begin{table}[htbp]
\centering
\caption{Validation performance of the deployed RF-DETR Small detector and the YOLOv8n-cls cover pose classifiers. Classifier values are final-epoch validation accuracies.}
\label{tab:technical-baseline}
\begin{tabular}{@{}llc@{}}
\toprule
Module & Metric & Result \\
\midrule
\multirow{3}{*}[-0.5em]{RF-DETR Small detection}
  & Validation precision  & 0.967 \\
  & Validation recall     & 0.870 \\
  & Validation mAP@0.5    & 0.949 \\
\midrule
\multirow{2}{*}[-0.3em]{YOLOv8n-cls pose classifier}
  & Front cover, final-epoch val.\ acc. & 96.6\% \\
  & Rear cover, final-epoch val.\ acc.  & 95.8\% \\
\bottomrule
\end{tabular}
\end{table}

RF-DETR Small achieved validation precision of 0.967, recall of 0.870, and mAP@0.5 of 0.949.
The YOLOv8n-cls front- and rear-cover pose classifiers achieved final-epoch validation accuracies of 96.6\% and 95.8\%, respectively. The released classifiers use the selected \texttt{best.pt} checkpoints; their checkpoint-specific validation metrics are reported in the model repositories.
The experimental piston-rod classifier showed substantially lower validation performance (top-1 accuracy of 28.5\% for the selected checkpoint and 24.5\% at the final epoch).
Its validation performance is reported separately
because the classifier was included in the experimental system
but was not validated to the same standard as the cover
classifiers. Consequently, the results of piston-rod-related
stage verification should be interpreted with this limitation
in mind.
Manual review of study logs recorded 11 false-negative events across 677 analyzed stage records: 7/207 under full guidance (3.38\%), 2/227 under mid guidance (0.88\%), and 2/243 under no guidance (0.82\%). These are intervention-log counts, not rates from an independently annotated trigger benchmark; true-positive and false-trigger rates are therefore not reported.

\subsection{Motion Quality: Primary Result \& Per-Metric Analysis}\label{subsec:primary-result}

The assistance effect was monotonic (LRT: $LR = 54.82$, $df = 2$, $p < 1.3 \times 10^{-12}$). 
Relative to full guidance, the estimated composite score was 0.27 points lower under mid guidance (95\% confidence interval (CI) $[-0.36, -0.19]$) and 0.42 points lower under no guidance (95\% CI $[-0.52, -0.33]$). These estimates and intervals describe differences in the four-metric composite's units, not standard deviations. The intervals suggest a consistent condition difference, although the sample included only eight participants.
Disassembly stages scored higher than assembly stages (+0.11 composite units, p = 0.003), and this covariate was included in all primary models.

As a sensitivity check for unequal stage counts across conditions, a trial-level model with one motion-quality observation per trial and \texttt{num\_stages} as an additional covariate yielded consistent condition coefficients (mid $\beta = -0.30$, no $\beta = -0.40$; both $p < 0.001$).

The model explained 19.1\% of the variance as fixed effects and 24.7\% including both participant and trial random intercepts (marginal $R^2 = 0.191$, conditional $R^2 = 0.247$). 
The participant-level variance component was 0.0140 and the trial-within-participant component was 0.0008. The latter was smaller than the participant component in this fitted model.
Leave-one-participant-out sensitivity analysis confirmed that the direction and significance of both effects were preserved across all eight refits (mid $\beta$ range $[-0.30, -0.25]$; no $\beta$ range $[-0.46, -0.40]$). 
Residual diagnostics showed 11 stage rows (1.6\%) with standardized residuals above 3 in absolute value, distributed across seven participants and all three conditions; refitting the model without these rows did not change the sign or significance of the condition effect.

The condition effect was clearest for smoothness and fragmentation (Table~\ref{tab:per-metric}). 
Relative to full guidance, jerk index was higher under mid ($\beta = +0.89$, log scale, $p < 0.001$) and no guidance ($\beta = +1.06$, $p < 0.001$); 
micro-stop counts were also higher (mid $\beta = +0.42$, no $\beta = +0.74$, both $p < 0.001$). 
Path inefficiency was higher under no guidance ($\beta = +0.15$, $p = 0.005$), but the mid contrast was not significant ($p = 0.53$). 
For pause ratio, the analysis found no evidence of a condition difference ($p = 0.27$); this does not establish that decision time was unchanged.

\begin{table}[htbp]
    \centering
    \caption{Per-metric linear mixed model (LMM) and count-model results. Omnibus statistics are likelihood-ratio tests for the mixed-effects models and a joint Wald test with participant-clustered standard errors for the negative-binomial model. Benjamini-Hochberg (BH) correction was applied across the four metrics.}
    \label{tab:per-metric}
    \small
    \setlength{\tabcolsep}{2pt}
    \begin{tabular}{@{}llcccc@{}}
        \toprule
        Metric & Model & Mid vs Full ($\beta$) & No vs Full ($\beta$) & \begin{tabular}[c]{@{}c@{}}Omnibus\\statistic\end{tabular} & \begin{tabular}[c]{@{}c@{}}BH-$p$\\(omnibus)\end{tabular} \\
        \midrule
        Jerk Index & LMM (log1p) & $+0.892$ & $+1.062$ & $88.26$ & $<0.001$ \\
        Micro-stops & Negative binomial & $+0.419$ & $+0.743$ & $59.56$ & $<0.001$ \\
        Path Inefficiency & LMM (log) & $+0.039$ & $+0.149$ & $7.87$ & $0.026$ \\
        Pause Ratio & LMM (logit) & $+0.144$ & $-0.095$ & $2.61$ & $0.272$ \\
        \botrule
    \end{tabular}
\end{table}

\subsection{Learning Trajectories}\label{subsec:learning-results}

\begin{figure}[htbp]
  \centering
   \includegraphics[width=0.58\linewidth]{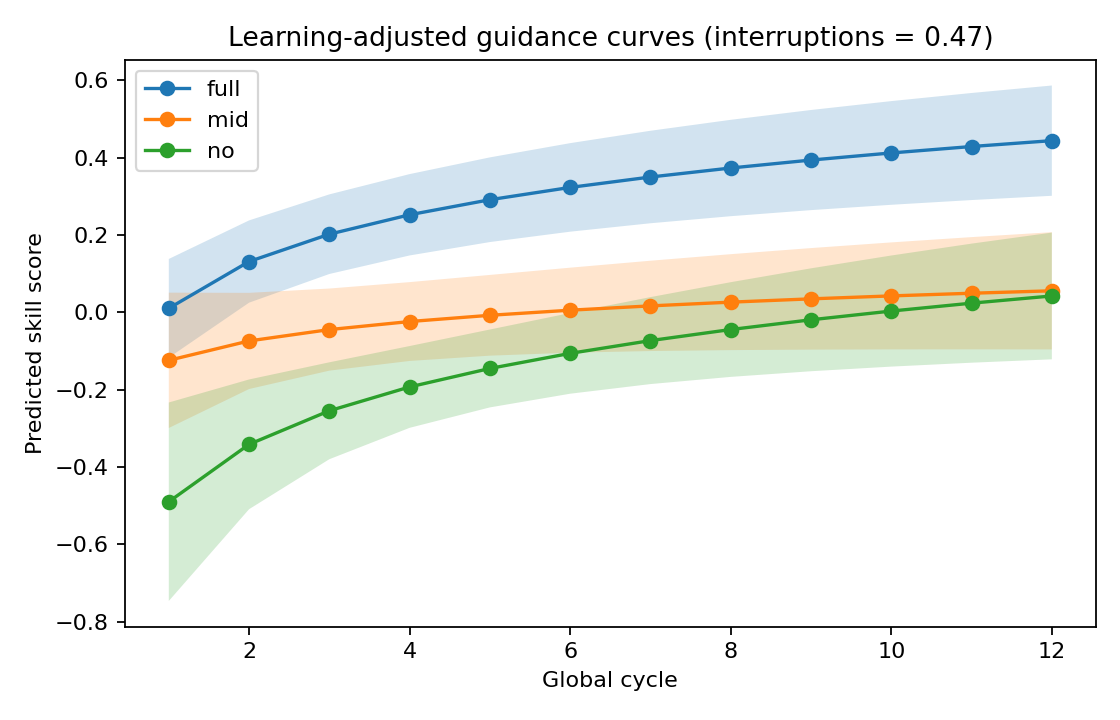}
  \caption{Model-estimated learning curves with 95\% confidence bands.
Interruptions are held at the dataset mean. The mid- and no-guidance
curves approach one another near cycle 12 but do not cross within
the observed range}
  \label{fig:learning}
\end{figure}

The joint condition-by-trial interaction was not significant in the primary model (interaction LRT: $LR = 3.26$, $df = 2$, $p = 0.20$), and Akaike information criterion (AIC) favored a logarithmic over a linear specification ($\Delta AIC = 6.4$). Individual slope estimates are reported descriptively. 

Under the log-cycle model, the full-guidance slope estimate was +0.17 score units per log-cycle. 
The mid-guidance slope was +0.07 (difference vs. full $\beta = -0.10$, $p = 0.11$) and was not distinguishable from zero ($p = 0.17$). 
The no-guidance slope was +0.21 (difference vs. full $\beta = +0.04$, $p = 0.61$); no guidance began from a lower estimated score ($-0.49$ at cycle 1). 
Because the model is logarithmic, all fitted curves flatten over time, and the small mid slope should not by itself be considered a plateau. 
Slope contrasts against zero are secondary; the joint interaction test above is the primary inferential statistic.
The fitted mid and no-guidance curves approach one another near the end of the observed range but do not cross by cycle 12 (Figure~\ref{fig:learning}). Their confidence bands overlap, and any later crossing would be extrapolation; it was not tested.

\subsection{Effect of Interruptions}\label{subsec:interruption-results}

Interruption frequency did not differ significantly across conditions (mean per stage: full $= 0.47$, mid $= 0.47$, no $= 0.48$). 
The raw interruption count was not a significant predictor of motion quality ($\beta = -0.006$, 95\% CI $[-0.055, +0.044]$, $p = 0.83$). 
To check whether this null result masked differential effects, we refitted the model replacing the raw count with specific components (operator errors, system false negatives, and clarification requests). 
None of these components predicted skill score ($p = 0.15$ for operator errors, $p = 0.64$ for false negatives, and $p = 0.39$ for clarification requests), and the condition effect remained unchanged and highly significant across all models (mid condition $\beta \approx -0.28$, $p < 10^{-9}$; no condition $\beta \approx -0.42$ to $-0.43$, $p < 10^{-18}$). Excluding interrupted trials gave similar estimates (mid $\beta=-0.256$, $p<10^{-9}$; no $\beta=-0.388$, $p<10^{-16}$).

\subsection{Task Completion and Errors}\label{subsec:completion-results}

In the participant-intercept model controlling for task type, log trial number, and detected stage count, there was no evidence of a condition effect on log completion time (mid $\beta=-0.022$, 95\% CI $[-0.136,0.091]$, $p=0.70$; no $\beta=-0.002$, 95\% CI $[-0.114,0.109]$, $p=0.97$). More detected stages were associated with longer duration ($\beta=0.053$, $p=0.001$), while duration declined over session experience ($\log\_global$ $\beta=-0.362$, $p<0.001$).

The Poisson model response was the number of unique failed stages per trial, counting a repeatedly failed stage once. It included condition, log trial number, and the number of FSM stage decisions in the trial as predictors, with participant-clustered robust standard errors; stage count was included as a covariate, not an exposure offset. There was no evidence of a condition difference (mid $p = 0.36$, no $p = 0.15$). Raw failed-decision totals were 11 under full, 9 under mid, and 9 under no guidance; these unadjusted totals do not establish that any condition reduced errors. 
The duration model included participant random intercepts. 
In a participant-level mixed-effects model of trial-averaged motion quality, adjusting for condition, log trial number, and task type, motion quality was negatively associated with log duration ($\beta=-0.22$, 95\% CI $[-0.38,-0.05]$, $p = 0.009$). 
This adjusted association is observational and is limited by eight participant clusters.

\subsection{Questionnaire Results}\label{subsec:questionnaire-results}

All eight participants completed four per-trial Likert items. 
Helpfulness differed by condition (Friedman $\chi^2 = 12.97$, $p = 0.0015$, Kendall's $W = 0.81$); full and mid guidance were rated above no guidance (both Holm-adjusted $p = 0.023$), with no detected full--mid difference ($p = 0.18$). Ease of understanding also differed ($\chi^2 = 12.60$, $p = 0.0018$, $W = 0.79$); all pairwise comparisons were significant (full--no $p_{Holm}=0.023$, mid--no $p_{Holm}=0.036$, full--mid $p_{Holm}=0.041$). For the two guidance-specific items, self-assessed independence averaged 3.90 under full and 3.81 under mid guidance, while reported reliance on instructions averaged 5.48 and 5.86, respectively. These items were not applicable under no guidance, so they were not compared across all three conditions.

In the debrief, five of eight participants chose mid guidance as best for learning and five chose full guidance for re-learning from scratch. For daily use, four preferred no guidance, three mid, and one full. Three participants each selected mid and no guidance as least distracting.

%%=============================================================%%
%% 6. DISCUSSION                                               %%
%%=============================================================%%
\section{Discussion}\label{sec:discussion}

\subsection{Interpretation of the Main Findings}\label{subsec:interpretation}

The primary model identified an association between guidance condition and the composite motion-quality score: mid and no guidance had lower estimated scores than full guidance by 0.27 and 0.42 points, respectively (Section~\ref{subsec:primary-result}). These are differences in the study's composite-score units, not standardized effects. Component analyses showed the clearest condition differences for jerk and micro-stop count; the analysis found no evidence of a condition difference in pause ratio. The reported models found no condition effect on completion time or failed-stage counts. Thus, the observed association concerns movement quality in this task; it does not establish faster completion or fewer errors.

The condition-by-trial interaction in the primary model was not significant ($LR=3.26$, $p=0.20$). The estimated slopes (+0.17 for full, +0.07 for mid, and +0.21 for no guidance per log-cycle) are therefore descriptive, and the apparent shallower mid-guidance trajectory is not evidence of a different learning rate or plateau. The logarithmic model also implies progressively smaller changes over cycles. The interruption-adjusted and interrupted-trial-exclusion analyses retained the condition association, but interruptions were not independently randomized and these checks do not establish that perception or trigger errors had no effect.

The pattern can be considered in relation to the motor-learning guidance hypothesis, which cautions that frequent augmented feedback may support performance during practice while potentially limiting retention or transfer if it substitutes for processing task-relevant information \citep{salmoni1984knowledge}. The higher estimated motion-quality score under full guidance is compatible with an online-performance benefit, but the study did not measure retention, transfer, cognitive load, or feedback dependence. Questionnaire preferences are self-reports and do not identify the mechanism behind the score differences. The findings therefore neither confirm nor refute the guidance hypothesis; they motivate a follow-up study that measures practice performance separately from retention and transfer.

\subsection{Design Rationale Revisited}\label{subsec:design-rationale}

The architecture uses a hybrid interaction trigger and discrete, task-specific orientation classes. On the validation split, the deployed detector achieved precision 0.967 and recall 0.870; the recall indicates that some target instances were missed in that image-level evaluation, but it does not quantify stage-level failures. The trigger requires regional motion together with a hand-position gate and temporal confirmation. This conjunction is intended to reduce reliance on either cue alone, but also creates operating-point trade-offs: thresholds that suppress transient activations can delay or miss short interactions, while permissive thresholds can admit unintended events. The study logs contain 11 false-negative event codes among 677 analyzed stage records (7/207 full, 2/227 mid, and 2/243 no guidance; Section~\ref{subsec:technical-evaluation}). These counts came from manual review of intervention logs, not frame-level annotation against an independent ground truth. They are evidence that missed events occurred, but cannot estimate sensitivity, false-trigger rate, or comparative trigger performance across conditions. The null interruption--motion-quality association is not a trigger-accuracy test.

For orientation, the system classifies a finite set of task-relevant states rather than estimating unconstrained 6-DoF pose. This choice matches the evaluated verification decisions and avoids reliance on depth measurements that were unreliable on the reflective parts in preliminary tests (Section~\ref{subsec:pose-classification}). Those tests were limited to four synthetic object crops and do not establish a general limit for depth-based pose estimation. Likewise, the reported classification accuracies above 95\% apply only to front- and rear-cover classifiers, not to every component or assembly condition.

\subsection{Effect of Assistance Level}\label{subsec:assistance-effect}

The fitted curves differed descriptively, but the primary model did not find a statistically significant condition-by-trial interaction.

\emph{Full guidance} produced the highest estimated motion-quality level. Its estimated slope was positive ($+0.17$ per log-cycle), but was not the steepest point estimate: the no-guidance slope was $+0.21$ and its interaction with full was not significant ($p=0.61$).

\emph{Mid guidance} had a lower estimated score than full; in the interaction model, the cycle-1 condition contrast was $\beta=-0.135$ (95\% CI $[-0.322,0.053]$, $p=0.16$). 
Its slope was $+0.07$ per log-cycle (95\% CI $[-0.030,0.175]$; $p=0.17$ against zero), and its slope contrast with full was $\beta=-0.10$ ($p=0.11$). The joint interaction test was not significant ($LR = 3.26$, $p = 0.20$).
The apparent flattening also reflects the logarithmic model form. One possible explanation is that the intermediate interface changes how operators use prompts, but compliance and internalization were not measured; this remains a hypothesis for future study.

\emph{No guidance} began at a lower estimated level, while its slope did not differ detectably from full ($\beta=+0.04$, $p=0.61$). The fitted mid and no-guidance curves approach one another near cycle 12 but do not cross within the observed range. This descriptive convergence does not show that mid guidance prevents later improvement.

The questionnaire responses do not resolve the descriptive slope pattern. Five of eight participants selected mid guidance as their preferred condition for learning, while five selected full guidance for re-learning from scratch. For daily use, four selected no guidance. These small-sample preferences describe stated choices; they do not measure learning, cognitive load, or performance and should not be interpreted as a ranking of the conditions.

\subsection{Comparison with Related Work}\label{subsec:comparison-related-work}

The present system differs from prior work in the level at which the design decisions are made. \citet{zhang2025stage} demonstrated accurate stage verification from multiple visual cues but did not close the loop back to operator guidance; \citet{zhou2024assembly} combined object detection with hand trajectories for procedure inspection, but without a configurable assistance layer. AR-based guidance systems \citep[e.g.,][]{wang2021augmented} deliver in-situ instructions without continuous verification of what the operator actually did, while action-recognition benchmarks such as Assembly101~\citep{sener2022assembly101} and AssemblyHands~\citep{ohkawa:cvpr23} evaluate recognition in isolation from verification and guidance.

The contribution of the present work is therefore integration rather than a new perception algorithm. 
Object detection, task-specific orientation classification, DTW-based trajectory verification, and two-level finite-state reasoning are combined into a single pipeline whose outputs drive a configurable assistance interface. 
Our study compares three assistance levels within participants over repeated trials and describes the shape of the fitted learning curves; the mid-guidance slope was shallower than full but the joint interaction was not significant with eight participants. 
Larger studies would be needed to confirm whether this pattern is systematic.

\subsection{Practical Implications}\label{subsec:practical-implications}

The results motivate further evaluation of configurable guidance, but do not identify an optimal condition for training or production. Guidance was assigned by study condition rather than selected from a validated estimate of operator expertise, and production use was not evaluated. The architecture makes the guidance interface separable from perception and procedural verification, which supports system-level adaptation as a design objective; this study did not test the effect of changing individual modules or selecting assistance online.

\subsection{Limitations}\label{subsec:limitations}

The evidence is limited by the small sample (eight participants, 70 valid trials), one assembly task, one fixed overhead camera, and a controlled laboratory setting. Participants were volunteers recruited at the host institution; the evaluation is not an in-production validation and does not establish generalization to other operators, tasks, workstations, or camera views. The condition-by-trial interaction was not significant, so the estimated slope differences are descriptive. Conditions alternated within participants, leaving possible learning carryover; later cycles also contain fewer participants because sessions stopped at different points (Section~\ref{subsec:user-study}). Participants sometimes repeated stages or ignored prompts. Each unsuccessful FSM decision was counted; in these records, decision-level and unique failed-stage counts matched because no stage was recorded as failing more than once within a trial. The study therefore characterizes outcomes under assigned guidance conditions, not the effect of consistent prompt compliance. Although the duration model included participant random intercepts and the failure model used participant-clustered standard errors, stage-level models did not include condition random slopes because their variance components were at the boundary in this sample.

The perception and interaction failure analysis is incomplete. Detector precision and recall were measured on the validation split, while the 11 stage-level false-negative codes were identified by manual review of intervention logs; no independently annotated event benchmark was available, and false-trigger rates were not measured. Consequently, the effect of missed or spurious triggers on trajectory segmentation and condition estimates remains uncertain. The implementation tolerates hand-tracking gaps of up to 10 frames and interpolates missing trajectory samples, but the study did not quantify the frequency, duration, or downstream impact of tracking dropouts. Object-detection inference runs on a remote server; end-to-end latency and its variability were not measured, so timing robustness under operational network or compute load is unknown. No component-wise ablation was conducted, preventing attribution of system-level outcomes to individual perception, reasoning, or interface modules. Finally, pose classification is task-specific, and the preliminary depth-based comparison was limited to a small set of synthetic crops.
The piston-rod pose classifier was preliminary and did not achieve validation performance comparable to the cover classifiers. Since it was available within the experimental system, its predictions may have affected the verification of piston-rod-related stages. The present study does not establish reliable piston-rod orientation classification.

\subsection{Future Work}\label{subsec:future-work}

Follow-up work should separate online performance from retention and transfer, recruit a larger and more diverse sample, and test additional assembly tasks, workstations, and camera views. Trigger evaluation should use independently annotated event-level ground truth and report missed and spurious events across operating conditions. Tracking dropout rates and end-to-end remote-inference latency should be measured alongside their effect on stage decisions and feedback timing. Component ablations would help quantify the contribution of the perception, interaction, procedural-reasoning, and guidance modules. Finally, a production study is needed to assess reliability and usability under sustained use and realistic operating constraints.

%%=============================================================%%
%% 7. CONCLUSION                                               %%
%%=============================================================%%
\section{Conclusion}\label{sec:conclusion}

Manual assembly in low-volume and high-variability production can benefit from computer-vision support while retaining the operator's role. This article presented a single-camera RGB framework combining object detection, task-specific pose classification, hand-trajectory comparison, two-level finite-state reasoning, and configurable visual guidance.

In a study of eight participants, full guidance had the highest estimated composite score. 
Mid and no guidance were lower by 0.27 and 0.42 composite points, respectively (95\% CIs $[-0.36,-0.19]$ and $[-0.52,-0.33]$; omnibus $LR=54.82$, $p<1.3\times10^{-12}$). These contrasts are composite-score units, not standard deviations. 
Estimated learning slopes were smallest under mid guidance and not distinguishable from zero, while no guidance had a slope similar to full from a lower starting level. The joint interaction test was not significant with this sample, so the shape differences are descriptive. 
The fitted mid and no-guidance curves approach one another near the end of the observed range but do not cross by cycle 12. Reported analyses found no condition effect on completion time or failed-stage counts; no speed or error-reduction benefit is claimed.

The results suggest that assistance level may relate to both motion-quality level and change over practice. The small within-participant sample and overlapping learning-curve intervals limit conclusions about trajectory shape. Larger studies with condition-specific learning tests and retention measures are needed before making deployment recommendations.

The study reports an integration of stage verification and configurable guidance using a single RGB camera. Selecting assistance according to operator expertise remains a design objective for future evaluation.

%%=============================================================%%
%% BACK MATTER                                                 %%
%%=============================================================%%
\section*{Acknowledgements}

The authors thank the volunteers who participated in the experimental evaluation. This work was carried out within the Inclu5ion project at Fraunhofer Italia Research.

\section*{Declarations}

\begin{itemize}
\item \textbf{Funding.} This work was carried out within the Inclu5ion project (Industry 5.0 Driven Design of Production Systems for Improving Inclusion and Equality of Disabled Workers), funded by the Autonomous Province of Bolzano (Research Südtirol/Alto Adige 2024, CUP: F53C23000850003).
\item \textbf{Competing interests.} The authors have no relevant financial or non-financial interests to disclose.
\item \textbf{Ethics approval.} Formal ethics committee approval was not required for this study, as all participants were internal volunteers from Fraunhofer Italia Research. All participants provided informed consent and data were anonymized. Participation was voluntary and uncompensated. The study involved no deception, invasive procedures, or sensitive personal data.
\item \textbf{Consent to participate.} Informed consent was obtained from all individual participants included in the study.
\item \textbf{Data availability.} The study metadata, intervention and questionnaire tables, statistical outputs, workbench homographies, and numerical reference trajectories are publicly available at \url{https://huggingface.co/datasets/fhi/piston-assembly-study}. The study recordings, detector and pose-classifier datasets, and trained model weights are provided in the linked repositories listed at \url{https://github.com/fraunhofer-italia/vision-assembly-assistance/blob/main/docs/artifacts.md}. The release includes 69 study recordings and metadata for 70 valid trials. The datasets are released under MIT; RF-DETR and YOLO model weights retain their Apache-2.0 and AGPL-3.0 licenses, respectively.
\item \textbf{Code availability.} The system implementation, task configurations, training utilities, and statistical analysis scripts are publicly available under the MIT License at \url{https://github.com/fraunhofer-italia/vision-assembly-assistance}.
\item \textbf{Author contributions.} Naveen Joseph Delphin: Conceptualization, Data curation, Formal analysis, Investigation, Methodology, Software, Validation, Visualization, Writing -- original draft; Leonardo Venturoso: Conceptualization, Methodology, Software, Supervision, Writing -- review \& editing; Isabella Soraruf: Conceptualization, Project administration, Resources, Supervision, Writing -- review \& editing; Marco Todescato: Conceptualization, Project administration, Resources, Supervision, Writing -- review \& editing.
\end{itemize}

%%=============================================================%%
%% BIBLIOGRAPHY                                               %%
%%=============================================================%%
\bibliography{references}

\end{document}